\documentclass[11pt,a4paper]{article}
\usepackage[hyperref]{emnlp-ijcnlp-2019}
\usepackage{times}
\usepackage{latexsym}

\usepackage{url}

\aclfinalcopy 

\usepackage{xcolor}
\usepackage{amsmath,amsthm}
\usepackage{amssymb,dsfont}
\usepackage{graphicx}
\usepackage{multirow}
\usepackage{todonotes}
\usepackage{booktabs,caption}
\usepackage[flushleft]{threeparttable}
\usepackage{textcomp}
\usepackage{subcaption}

\newcommand{\ignore}[1]{}

\newcommand{\qiangchange}[1]{{{#1}}}
\newcommand{\sanjaychange}[1]{{{#1}}}

\newcommand{\temprel}[1]{{\em #1}}
\newcommand{\ourkb}{\textsc{TemProb}}
\usepackage{balance}
\title{
An Improved Neural Baseline for Temporal Relation Extraction
}
\author{Qiang Ning,$^{1}$~Sanjay Subramanian,$^{2}$~and~Dan Roth$^{1,2}$\\
$^{1}$University of Illinois at Urbana-Champaign\\
$^{2}$University of Pennsylvania\\
{\tt qning2@illinois.edu}, {\tt \{subs,danroth\}@seas.upenn.edu} }

\date{}

\begin{document}
\maketitle
\begin{abstract}
Determining temporal relations (e.g., \temprel{before} or \temprel{after}) between events has been a challenging natural language understanding task, partly due to the difficulty to generate large amounts of high-quality training data. Consequently, neural approaches have not been widely used on it, or showed only moderate improvements. This paper proposes a new neural system that achieves about 10\% absolute improvement in accuracy over the previous best system (25\% error reduction) on two benchmark datasets. The proposed system is trained on the state-of-the-art MATRES dataset and applies contextualized word embeddings, a Siamese encoder of a temporal common sense knowledge base, and global inference via integer linear programming (ILP). We suggest that the new approach could serve as a strong baseline for future research in this area.
\end{abstract}

\section{Introduction}
\label{sec:intro}
Temporal relation (TempRel) extraction has been considered as a major component of understanding time in natural language \cite{DoLuRo12,ULADVP13,MSAAEMRUK15,LCUMAP15,NFWR18}.
However, the annotation process for TempRels is known to be time consuming and difficult even for humans, and existing datasets are usually small and/or have low inter-annotator agreements (IAA); e.g., \citet{ULADVP13,CCMB14,GormanWrPa16} reported Kohen's $\kappa$ and $F_1$ in the 60's.
\qiangchange{Albeit the significant progress in deep learning nowadays,} neural approaches have not been used extensively for this task, or showed only moderate improvements \cite{DMLBS17,LMDBS17,MengRu18}.
\qiangchange{We think it is important for to understand: is it because we missed a ``magic'' neural architecture, because the training dataset is small, or because the quality of the dataset should be improved?}

\qiangchange{Recently, \citet{NingWuRo18} introduced a new dataset called {\em Multi-Axis Temporal RElations for Start-points} (MATRES). MATRES is still relatively small in its size (15K TempRels), but has a higher annotation quality from its improved task definition and annotation guideline.}
This paper uses MATRES to show that a long short-term memory (LSTM) \cite{HochreiterSc97} system can readily outperform the previous state-of-the-art system, CogCompTime \cite{NZFPR18}, by a large margin.
\qiangchange{The fact that a standard LSTM system can significantly improve over a feature-based system on MATRES indicates that neural approaches have been mainly dwarfed by the quality of annotation, instead of specific neural architectures or the small size of data.}

\qiangchange{To gain a better understanding of the standard LSTM method,} we extensively compare the usage of various word embedding techniques, including word2vec \cite{MCCD13c}, GloVe \cite{PenningtonSoMa14}, FastText \cite{BGJM16}, ELMo \cite{PNIGCLZ18}, and BERT \cite{DCLT18}, and show their impact on TempRel extraction.
Moreover, we further improve the LSTM system by injecting knowledge from an updated version of \ourkb{}, an automatically induced temporal common sense knowledge base that provides {\em typical} TempRels between events\footnote{For example, ``explode" typically happens {\em before} ``die".}~\cite{NWPR18}.
Altogether, these components improve over CogCompTime by about 10\% in $F_1$ and accuracy. The proposed system is public\footnote{\url{https://cogcomp.org/page/publication_view/879}} and can serve as a strong baseline for future research.
\section{Related Work}
\label{sec:related}


Early computational attempts to TempRel extraction include \citet{MVWLP06,ChambersWaJu07,BethardMaKl07,VerhagenPu08}, which aimed at building classic learning algorithms (e.g., perceptron, SVM, and logistic regression) using {\em hand-engineered features} extracted for each pair of events.
The frontier was later pushed forward through continuous efforts in a series of SemEval workshops \cite{VGSHKP07,VSCP10,ULADVP13,BDSPV15,BSCDPV16,BSPP17}, and significant progresses were made in terms of data annotation \cite{SBFPPG14,CMCB14,MGCAV16,GormanWrPa16}, structured inference \cite{ChambersJu08b,DoLuRo12,CCMB14,NFWR18}, and structured machine learning \cite{YRAM09,LeeuwenbergMo17,NingFeRo17}.

Since TempRel is a specific relation type, it is natural to borrow recent neural relation extraction approaches \cite{ZLLZZ14,ZZHY15,ZhangWa15,XuHuDe16}.
There have indeed been such attempts, e.g., in clinical narratives \cite{DMLBS17,LMDBS17,TFNT17} and in newswire \cite{ChengMi17,MengRu18,LeeuwenbergMo18}.
However, their improvements over feature-based methods were moderate (\citet{LMDBS17} even showed negative results). 
Given the low IAAs in those datasets, it was unclear whether it was simply due to the low data quality or neural methods inherently do not work well for this task.

A recent annotation scheme, \citet{NingWuRo18}, introduced the notion of multi-axis to represent the temporal structure of text, and identified that one of the sources of confusions in human annotation is asking annotators for TempRels across different axes. When annotating only same-axis TempRels, along with some other improvements to the annotation guidelines, MATRES was able to achieve much higher IAAs.\footnote{Between experts: Kohen's $\kappa\approx0.84$. Among crowdsourcers: accuracy 88\%. More details in \citet{NingWuRo18}.}
This dataset opens up opportunities to study neural methods for this problem.
In Sec.~\ref{sec:system}, we will explain our proposed LSTM system, and also highlight the major differences from previous neural attempts.

\section{Neural TempRel Extraction}
\label{sec:system}

\begin{figure*}[htbp!]
	\centering
	\includegraphics[width=\textwidth]{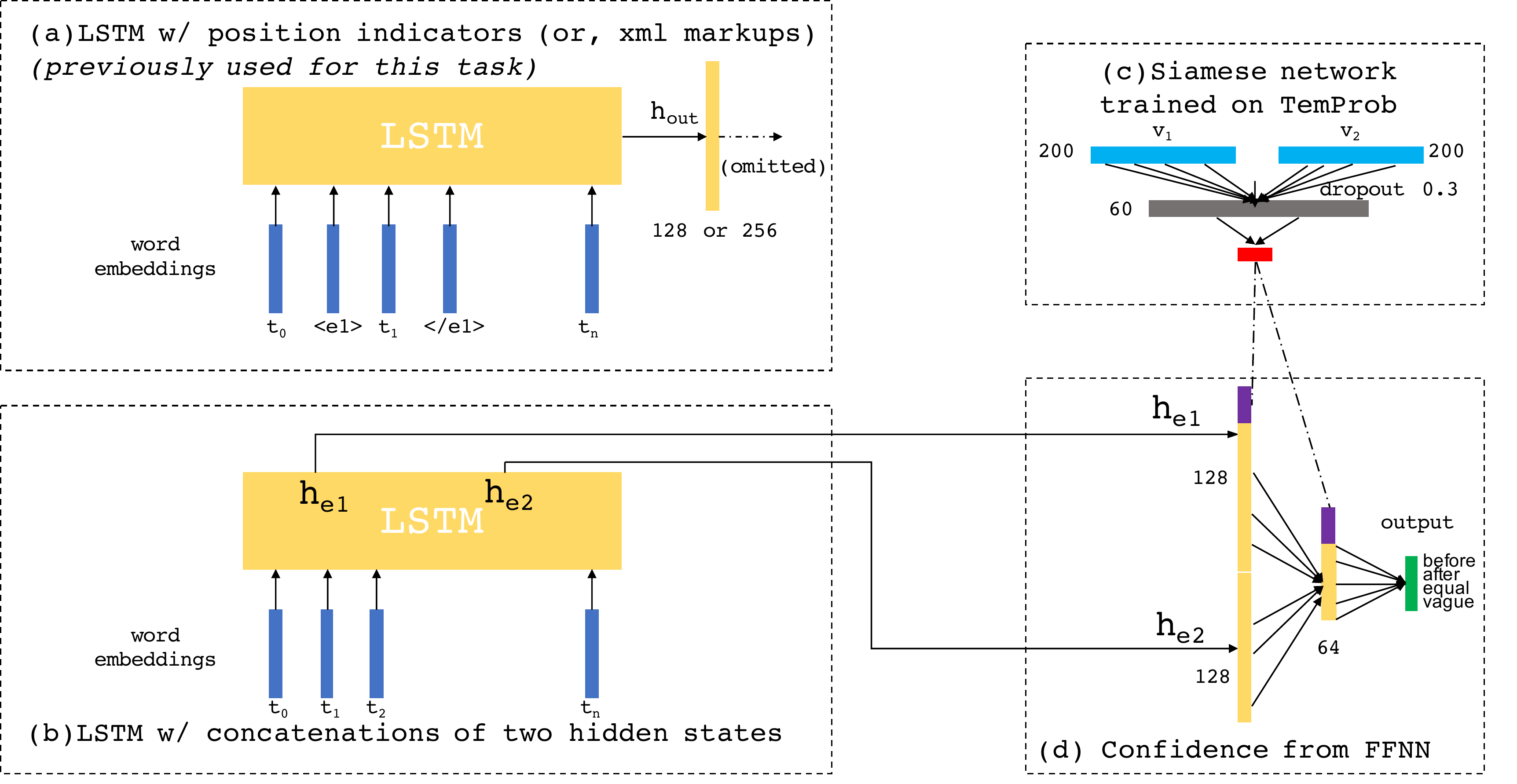}
	\caption{Overview of the neural network structures studied in this paper. Networks (a) and (b) are two ways to handle event positions in LSTMs (Sec.~\ref{subsec:position}). (c) The Siamese network used to fit TemProb (Sec.~\ref{subsec:common sense}). \qiangchange{Once trained on TemProb, the Siamese network is fixed when training other parts of the system.} (d) The FFNN that generates confidence scores for each label. Sizes of hidden layers are already noted. Embeddings of the same color share the same matrix.}
	\label{fig:NN}
\end{figure*}

One major disadvantage of feature-based systems is that errors occurred in feature extraction propagate to subsequent modules.
Here we study the usage of LSTM networks\footnote{We also tried convolutional neural networks but \sanjaychange{did not observe that CNNs improved performance significantly compared to the LSTMs}. Comparison between LSTM and CNN is also not the focus of this paper.} on the TempRel extraction problem as an end-to-end approach that only takes a sequence of word embeddings as input (assuming that the position of events are known).
Conceptually, we need to feed those word embeddings to LSTMs and obtain a vector representation for a particular pair of events,
which is followed by a fully-connected, feed-forward neural network (FFNN) to generate confidence scores for each output label.
Based on the confidence scores, global inference is performed via integer linear programming (ILP), which is a standard procedure used in many existing works to enforce the transitivity property of time \cite{ChambersJu08,DoLuRo12,NingFeRo17}.
An overview of the proposed network structure and corresponding parameters can be found in Fig.~\ref{fig:NN}. Below we also explain the main components.

\subsection{Handling Event Positions}
\label{subsec:position}
Each TempRel is associated with two events, and for the same text, different pairs of events possess different relations, so it is critical to indicate the positions of those events when we train LSTMs for the task.
The most straightforward way is to concatenate the hidden states from both time steps that correspond to the location of those events (Fig.~\ref{fig:NN}b).
\citet{DMLBS17} handled this issue differently, by adding XML tags immediately before and after each event (Fig.~\ref{fig:NN}a).
For example, in the sentence, {\em After \textbf{eating} dinner, he \textbf{slept} comfortably}, where the two events are bold-faced, they will convert the sequence into {\em After $<$e1$>$ eating $<$/e1$>$ dinner, he $<$e2$>$ slept $<$/e2$>$ comfortably.}
The XML markups, which was initially proposed under the name of {\em position indicators} for relation extraction \cite{ZhangWa15}, uniquely indicate the event positions to LSTM, such that the final output of LSTM can be used as a representation of those events and their context.
We compare both methods in this paper, and as we show later, the straightforward concatenation method is already as good as XML tags for this task.

\subsection{Common Sense Encoder (CSE)}
\label{subsec:common sense}
In naturally occurring text that expresses TempRels, connective words such as {\em since}, {\em when}, or {\em until} are often not explicit; nevertheless, humans can still infer the TempRels using common sense with respect to the events. For example, even without context, we know that {\em die} is typically after {\em explode} and {\em schedule} typically before {\em attend}.
\citet{NWPR18} was an initial attempt to acquire such knowledge, by aggregating automatically extracted TempRels from a large corpus. The resulting knowledge base, \ourkb{}, contains observed frequencies of tuples $(v1,v2,r)$ representing the probability of verb 1 and verb 2 having relation $r$ and it was shown a useful resource for TempRel extraction.

However, \ourkb{} is a simple counting model and fails (or is unreliable) for unseen (or rare) tuples.
For example, we may see (ambush, die) less frequently than (attack, die) in a corpus, and the observed frequency of (ambush, die) being \temprel{before} or \temprel{after} is thus less reliable. However, since ``ambush'' is semantically similar to ``attack'', the statistics of (attack, die) can actually serve as an auxiliary signal to (ambush, die).
Motivated by this idea, we introduce common sense encoder (CSE): We fit an updated version of \ourkb{} via a Siamese network \cite{BGLSS94} that generalizes to unseen tuples through the resulting embeddings for each verb (Fig.~\ref{fig:NN}c).
\qiangchange{Note that the \ourkb{} we use is reconstructed using the same method described in \citet{NWPR18} with the base method changed to CogCompTime. Once trained, CSE will remain fixed when training the LSTM part (Fig.~\ref{fig:NN}a or b) and the feedforward neural network part (Fig.~\ref{fig:NN}d). We only use CSE for its output. In the beginning, we tried to directly use the output (i.e., a scalar) and the influence on performance was negligible. 
Therefore, here we discretize the CSE output, change it to categorical embeddings, concatenate them with the LSTM output, and then produce the confidence scores (Fig.~\ref{fig:NN}d).} 
\section{Experiments}
\label{sec:experiments}
\subsection{Data}
The MATRES dataset\footnote{\url{http://cogcomp.org/page/publication_view/834}} contains 275 news articles from the TempEval3 workshop \cite{ULADVP13} with newly annotated events and TempRels.
It has 3 sections: TimeBank (TB), AQUAINT (AQ), and Platinum (PT).
We followed the official split (i.e., TB+AQ for training and PT for testing), and further set aside 20\% of the training data as the development set to tune learning rates and epochs.
We also show our performance on another dataset, TCR\footnote{\url{http://cogcomp.org/page/publication_view/835}} \cite{NFWR18}, which contains both temporal and causal relations and we only need the temporal part.
The label set for both datasets are \temprel{before}, \temprel{after}, \temprel{equal}, and \temprel{vague}.

\begin{table}[htbp!]
\centering\small
\begin{tabular}{c|c|c|c|c}
    \hline
            &	Purpose	&   \#Doc   &   \#Events    &   \#TempRels  \\
    \hline
    TB+AQ   &	Train	&   255     &   8K          &   13K         \\
    PT    	&	Test	&   20      &   537         &   837         \\
    TCR		&	Test	&	25		&	1.3K		&	2.6K		\\
    \hline
\end{tabular}
\caption{TimeBank (TB), AQUAINT (AQ), and Platinum (PT) are from MATRES \cite{NingWuRo18} and TCR from \citet{NFWR18}.}
\label{tab:stats}
\end{table}

\subsection{Results and Discussion}

We compare with the most recent version of CogCompTime, the state-of-the-art on MATRES.\footnote{\url{http://cogcomp.org/page/publication_view/844}}
Note that in Table~\ref{tab:position}, CogCompTime performed slightly different to \citet{NZFPR18}: CogCompTime reportedly had $F_1$=65.9 (Table~2 Line~3 therein) and here we obtained $F_1$=66.6.
In addition, \citet{NZFPR18} only reported $F_1$ scores, while we also use another two metrics for a more thorough comparison: classification accuracy (acc.) and temporal awareness $F_{\text{aware}}$, where the awareness score is for the graphs represented by a group of related TempRels (more details in the appendix).
We also report the average of those three metrics in our experiments.

Table~\ref{tab:position} compares the two different ways to handle event positions discussed in Sec.~\ref{subsec:position}: position indicators (P.I.) and simple concatenation (Concat), both of which are followed by network (d) in Fig.~\ref{fig:NN} (i.e., without using Siamese yet).
We extensively studied the usage of various pretrained word embeddings, including conventional embeddings (i.e., the medium versions of word2vec, GloVe, and FastText provided in the Magnitude package \cite{PSCA18}) and contextualized embeddings (i.e., the original ELMo and large uncased BERT, respectively); except for the input embeddings, we kept all other parameters the same.
We used cross-entropy loss and the StepLR optimizer in PyTorch that decays the learning rate by 0.5 every 10 epochs (performance not sensitive to it).

Comparing to the previously used P.I. \cite{DMLBS17}, we find that, with only two exceptions (underlined \qiangchange{in Table~\ref{tab:position}}), the Concat system saw consistent gains under various embeddings and metrics.
In addition, contextualized embeddings (ELMo and BERT) expectedly improved over the conventional ones significantly, although no statistical significance were observed between using ELMo or BERT (more significance tests in Appendix).

\begin{table}[htbp!]
	\centering\small
	\begin{tabular}{ lccccc } 
		\hline
		\textit{System} & \textit{Emb.} & \textit{Acc.} & $F_1$ & $F_{\text{aware}}$	&	Avg.\\
		\cmidrule(lr){1-6} 
		\multirow{5}{*}{P.I.} 
		& word2vec & 63.2 & 67.6 & \underline{60.5} & 63.8 \\ 
		& GloVe & 64.5 & 69.0 & \underline{61.1} & 64.9\\ 
		& FastText & 60.5 & 64.7 & 59.5 & 61.6\\ \cmidrule(lr){2-6} 
		& ELMo & 67.5 & 73.9 & 63.0 & \textbf{68.1}\\
		& BERT & 68.8 & 73.6 & 61.7 & 68.0\\
		\cmidrule(lr){1-6} 
		\multirow{5}{*}{Concat} 
		& word2vec & 65.0 & 69.5 & 59.4 & 64.6\\ 
		& GloVe & 64.9 & 69.5 & 60.9 & 65.1\\ 
		& FastText & 64.0 & 68.6 & 60.1 & 64.2\\  \cmidrule(lr){2-6} 
		& ELMo & 67.7 & 74.0 & 63.3 & 68.3\\
		& BERT & 69.1 & 74.4 & 63.7 & \textbf{69.1}\\
		\cmidrule(lr){1-6} 
		\multirow{2}{*}{Concat+CSE} 
		& ELMo & 71.7  & 76.7  & 66.0 & \textbf{71.5} \\
		& BERT & 71.3 & 76.3 & 66.5 & 71.4 \\
		\cmidrule(lr){1-6} 
		\multicolumn{2}{l}{CogCompTime\quad\quad\em\em -} & 61.6 & 66.6 & 60.8 & \textbf{63.0}\\
		\hline
	\end{tabular}
	\caption{Performances on the MATRES test set (i.e., \qiangchange{the PT section}). CogCompTime \cite{NZFPR18} is the previous state-of-the-art feature-based system. Position indicator (P.I.) and concatenation (Concat) are two ways to handle event positions in LSTMs (Sec.~\ref{subsec:position}). Concat+CSE achieves significant improvement over CogCompTime on MATRES.}
	\label{tab:position}
\end{table}

\ignore{
\begin{table}[htbp!]
	\centering\small
	\begin{tabular}{ lccccc } 
		\hline
		\textit{System} & \textit{Emb.} & \textit{Acc.} & $F_1$ & $F_{\text{aware}}$	&	Avg.\\
		\cmidrule(lr){1-6} 
		\multirow{5}{*}{P.I.} 
		& word2vec & 63.2 & 67.6 & \underline{60.5} & 63.8 \\ 
		& GloVe & 64.5 & 69.0 & \underline{61.1} & 64.9\\ 
		& FastText & 60.5 & 64.7 & 59.5 & 61.6\\ \cmidrule(lr){2-6} 
		& ELMo & 67.5 & 73.9 & 63.0 & \textbf{68.1}\\
		& BERT & 68.8 & 73.6 & 61.7 & 68.0\\
		\cmidrule(lr){1-6} 
		\multirow{5}{*}{Concat} 
		& word2vec & 65.0 & 69.5 & 59.4 & 64.6\\ 
		& GloVe & 64.9 & 69.5 & 60.9 & 65.1\\ 
		& FastText & 64.0 & 68.6 & 60.1 & 64.2\\  \cmidrule(lr){2-6} 
		& ELMo & 67.7 & 73.9 & 63.3 & 68.3\\
		& BERT & 69.1 & 74.4 & 63.7 & \textbf{69.1}\\
		\hline
	\end{tabular}
	\caption{Performances on the MATRES test set (i.e., PT). CogCompTime is the previous state-of-the-art, feature-based system. Position indicator (P.I.) and concatenation (Concat) are two ways to handle event positions in LSTMs (Sec.~\ref{subsec:position}).}
	\label{tab:position}
\end{table}
}

Given the above two observations, we further incorporated our common sense encoder (CSE) into ``Concat'' with ELMo and BERT in Table~\ref{tab:position}.
We split \ourkb{} into train (80\%) and validation (20\%). The proposed Siamese network (Fig.~\ref{fig:NN}c) 
was trained by minimizing the cross-entropy loss using Adam \cite{KingmaBa14} (learning rate 1e-4, 20 epochs, and batch size 500).  
We first see that CSE improved on top of Concat for both ELMo and BERT under all metrics,  confirming the benefit of \ourkb{}; second, as compared to CogCompTime, the proposed Concat+CSE achieved about 10\% absolute gains in accuracy and $F_1$, 5\% in awareness score $F_{\text{aware}}$, and 8\% in the three-metric-average metric, with $p<0.001$ per the McNemar's test.  Roughly speaking, the 8\% gain is contributed by LSTMs for 2\%, contextualized embeddings for 4\%, and CSE for 2\%.
Again, no statistical significance were observed between using ELMo and BERT.
Table~\ref{tab:cse} furthermore applies CogCompTime and the proposed Concat+CSE system on a different test set called TCR \cite{NFWR18}. Both systems achieved better scores (suggesting that TCR is easier than MATRES), while the proposed system still outperformed CogCompTime by roughly 8\% under the three-metric-average metric, consistent with our improvement on MATRES.

\begin{table}[htbp!]
	\centering\small
	\begin{tabular}{ lccccc } 
		\hline
		\textit{System} & \textit{Emb.} & \textit{Acc.} & $F_1$ & $F_{\text{aware}}$ & Avg.\\
		\cmidrule(lr){1-2} \cmidrule(lr){3-3} \cmidrule(lr){4-4} \cmidrule(lr){5-5} \cmidrule(lr){6-6}
		\multicolumn{2}{l}{CogCompTime~~~~~~-} & 68.1 & 70.7 & 61.6 & \textbf{66.8}\\
		\cmidrule(lr){1-6} 
		\multirow{2}{*}{Concat+CSE} 
		& ELMo & 80.8  & 78.6  & 69.9 & \textbf{76.4} \\
		& BERT & 78.4 & 77.0 & 69.0 & 74.9 \\
		\hline
	\end{tabular}
	\caption{Further evaluation of the proposed system, i.e., Concat (Table~\ref{subsec:position}) plus CSE (Sec.~\ref{subsec:common sense}), on the TCR dataset \cite{NFWR18}.}
	\label{tab:cse}
\end{table}

\section{Conclusion}
\label{sec:conclusion}

Temporal relation extraction has long been an important yet challenging task in natural language processing.
Lack of high-quality data and difficulty in the learning problem defined by previous annotation schemes inhibited performance of neural-based approaches.
The discoveries that LSTMs readily improve the feature-based state-of-the-art CogCompTime on the MATRES and TCR datasets by a large margin not only give the community a strong baseline, but also indicate that the learning problem is probably better defined by MATRES and TCR.
Therefore, we should move along that direction to collect more high-quality data, which can facilitate more advanced learning algorithms in the future.

\section*{Acknowledgements}
\qiangchange{
This research is supported by a grant from the Allen Institute for Artificial Intelligence (allenai.org),  the IBM-ILLINOIS Center for Cognitive Computing Systems Research (C3SR) - a research collaboration as part of the IBM AI Horizons Network, and contract HR0011-18-2-0052 with the US Defense Advanced Research Projects Agency (DARPA). Approved for Public Release, Distribution Unlimited.
The views expressed are those of the authors and do not reflect the official policy or position of the Department of Defense or the U.S. Government.
}
\appendix


\section{Illustration of the metrics used in this paper}
\label{subsec:metrics}

\begin{figure}[htbp!]
	\centering
	\includegraphics[width=.4\textwidth]{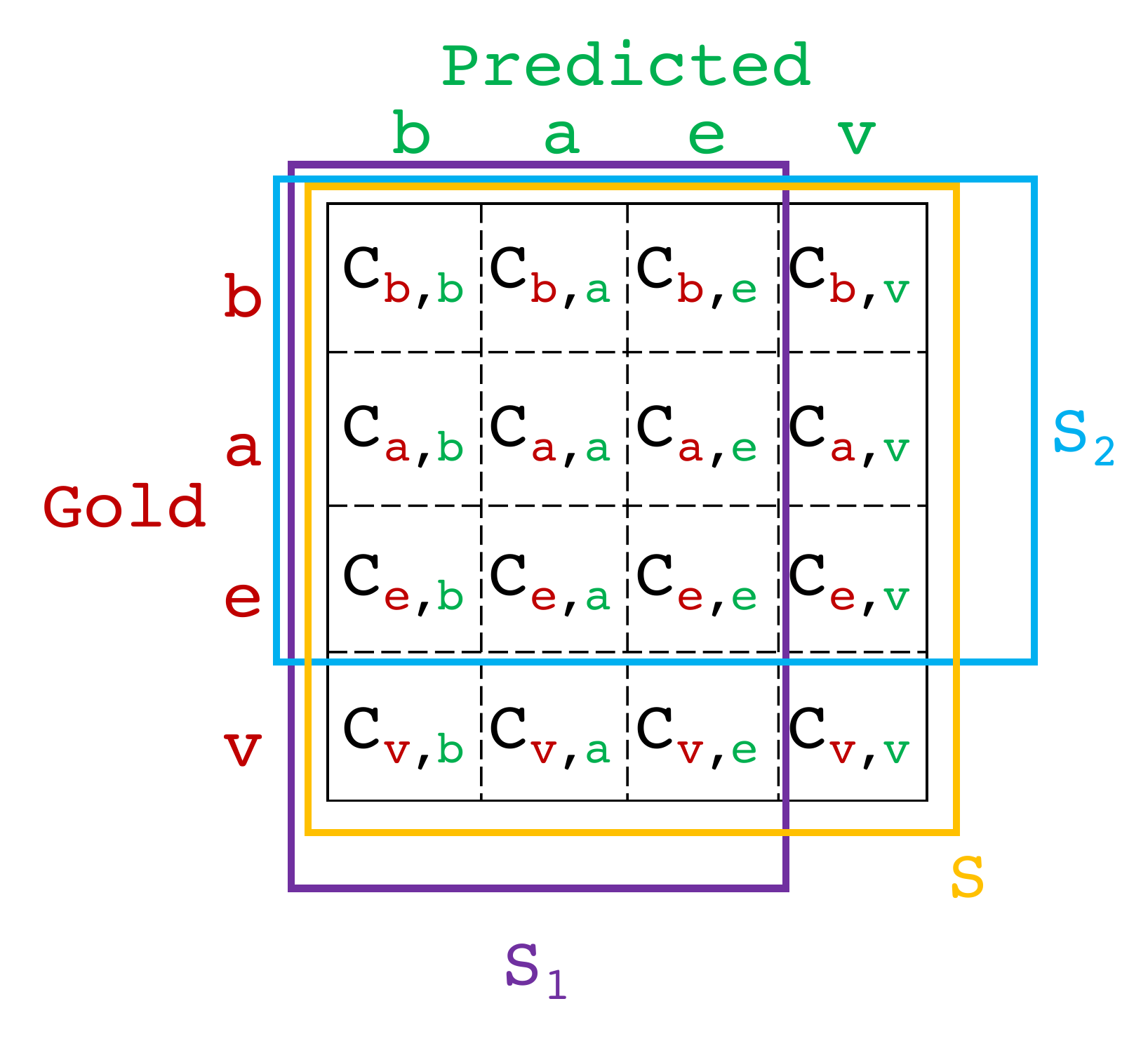}
	\caption{An example confusion matrix, where the four labels are \temprel{before} (b), \temprel{after} (a), \temprel{equal} (e), and \temprel{vague} (v), respectively. The variables, $S$, $S_1$, and $S_2$, are the summation of all the numbers in the corresponding area. (This figure is better viewed in color)}
	\label{fig:confusion}
\end{figure}

There are three widely-used evaluation metrics for the TempRel extraction task. 
The first standard metric is classification {\em accuracy} (Acc).
The second metric is to view this task as a general relation extraction task, treat the label of \temprel{vague} for TempRel as {\em no relation}, and then compute the precision and recall. 
The third metric came into use since the TempEval3 workshop \cite{ULADVP13}, which involves graph closure and reduction on top of the second metric, hoping to better capture how useful a TempRel system is..
Take the confusion matrix in Fig.~\ref{fig:confusion} for example. The three metrics used in this paper are
\begin{enumerate}
    \item Accuracy. $Acc=(C_{b,b}+C_{a,a}+C_{e,e}+C_{v,v})/S$.
    \item Precision, recall, and $F_1$. $P=(C_{b,b}+C_{a,a}+C_{e,e})/S_1$, $R=(C_{b,b}+C_{a,a}+C_{e,e})/S_2$, and $F_1=2PR/(P+R)$.
    \item Awareness score $F_{\text{aware}}$. Before calculating precision, perform a graph closure on the gold temporal graph and a graph reduction on the predicted temporal graph. Similarly, before calculating recall, perform a graph reduction on the gold temporal graph and a graph closure on the predicted temporal graph. Finally, compute the $F_1$ score based on this revised precision and recall. Since graph reduction and closure are involved in computing this metric, the temporal graphs all need to satisfy the global transitivity constraints of temporal relations (e.g., if A happened \temprel{before} B, and B happened \temprel{before} C, then C cannot be \temprel{before} A).
\end{enumerate}
In this paper, we also report the average of the three metrics above, which we call {\em three-metric-average}.

\section{Significance test for Tables~\ref{tab:position}-\ref{tab:cse}}
\label{subsec:significance}

In Table~\ref{tab:position}, we mainly compared the performance of position indicator (P.I.) and simple concatenation (Concat), using 5 different word embeddings and 3 metrics, so there were 15 performances for both P.I. and Concat. Under the paired t-test, Concat is significantly better than P.I. with $p<0.01$.

Another observation we had in Table~\ref{tab:position} was that contextualized embeddings, i.e., ELMo and BERT, were much better than conventional ones, i.e., word2vec, GloVe and FastText. For both P.I. and Concat, we found that the difference between contextualized embeddings and conventional embeddings was significant with $p<0.001$ under the McNemar's test \cite{Everitt92,Dietterich98b}; however, between the two contextualized embeddings, ELMo and BERT, we did not see a significant difference, although it has been reported that in many {\em other} tasks, that BERT is better than ELMo.

In Table~\ref{tab:cse}, we further improved Concat using the proposed common sense encoder (CSE). Under the McNemar's test, Concat+CSE was significantly better than Concat with $p<0.001$, no matter either ELMo or BERT was used. Again, no significant difference was observed between ELMo and BERT. Finally, since Concat+CSE improved over CogCompTime by a large margin either on MATRES or on TCR, it was not surprising to see that the proposed Concat+CSE is significantly better than CogCompTime with $p<0.001$ as well.
\bibliography{ccg-long,cited-long,new}
\bibliographystyle{acl_natbib}

\end{document}